\documentclass[runningheads]{llncs}

\usepackage{bm}        
\usepackage{amsmath}
\usepackage{amssymb}
\usepackage{tikz}
\usepackage{hyperref}
\usepackage{url}
\usetikzlibrary{positioning, arrows.meta}
\usepackage[T1]{fontenc}
\usepackage{graphicx,verbatim}
\usepackage{booktabs}
\usepackage{threeparttable}

\begin{document}

\title{Gradient-Based Latent Decomposition Reveals Mechanisms of Feature Degradation in Weakly Supervised Mammography}
\titlerunning{Gradient-Based Latent Decomposition in Weakly Supervised Mammography}

 \author{Vinceline Bertrand \and Ionut Cardei}
 \authorrunning{V. Bertrand and I. Cardei}
 \institute{Florida Atlantic University, Boca Raton, FL, USA \\ \email{\{vbertrand2023, icardei\}@fau.edu}}

\maketitle

\begin{abstract}
Weakly supervised hierarchical models exhibit a persistent asymmetry: coarse lesion-type features are preserved under reconstruction while fine-grained malignancy cues degrade---a pattern with direct consequences for the clinical reliability of breast cancer screening pipelines. We introduce gradient-based orthogonal latent decomposition for hierarchical Variational Autoencoders~(H-VAEs) to mechanistically explain this asymmetry. The latent space is partitioned into a task-aligned component~($z_1$), shaped by coarse supervisory gradients, and an orthogonal residual~($z_{\text{res}}$) capturing remaining representational capacity. On~3,550 mammographic Regions of Interest~(ROIs) from CBIS-DDSM, only~$\sim$4.4\% of latent magnitude aligns with supervisory gradients, leaving~$\sim$95.6\% in the orthogonal residual upon which fine-grained pathology prediction primarily depends. The model achieves Stage-1~AUC~0.866 and Stage 2~AUC~0.552, with a reconstruction stability gap of $\Delta_{\text{diag}}=5\%$ ($p=0.005$) and a classification gap of $\Delta_{\text{AUC}}=0.314$ ($p{<}0.001$). Latent ablation confirms that features for both tasks reside heavily in~$z_{\text{res}}$, structurally explaining why reconstruction degrades pathology stability disproportionately. Comparisons with Multi-Instance Learning~(MIL) and Multi-Task Learning~(MTL) confirm generalization across architectures and modalities. These findings reveal that in high-dimensional spaces, a single coarse supervisory signal isolates only a sparse 1D latent direction, forcing critical fine-grained features into the vulnerable residual subspace.
\end{abstract}

\keywords{Weakly supervised learning \and Variational autoencoder \and Latent decomposition \and Hierarchical classification \and Breast cancer screening}

\section{Introduction}

Breast cancer is the most commonly diagnosed cancer among women worldwide~\cite{siegel2023cancer}, and its early detection yields five-year survival rates exceeding~99\% for localized disease~\cite{society2025survival}. Despite promising performance, deep learning models remain difficult to integrate reliably into clinical workflows due to limited generalizability~\cite{zech2018variable}.

Annotation costs constrain many clinical applications: pixel-level labels for lesion segmentation are expensive and time-consuming, motivating Weakly Supervised Learning methods~\cite{zhou2016learning}. Convolutional classifiers trained on Region of Interest~(ROI) crops have achieved competitive Area Under the Receiver Operating Characteristic Curve~(AUC) on mammography benchmarks~\cite{shen2019deep}, yet these aggregate metrics can mask clinically meaningful information loss. Prior work has documented that models relying on weak supervision tend to preserve coarse structural features such as lesion type, while degrading under perturbation for fine-grained pathology assessment~\cite{rudin2019stop,raghu2019transfusion,icml_short}.

More broadly, prior work shows that without strong inductive biases, weak supervision does not fully constrain the latent space in generative models, limiting identifiability of task-relevant structure~\cite{locatello2020weakly}. In other words, the problem is underconstrained: many different latent configurations can produce the same coarse label while encoding different fine-grained details. As a result, pathology-relevant features are not required to align with the supervised direction and may remain weakly organized in latent space.This motivates a geometric analysis of how supervision shapes latent robustness for fine-grained prediction. Understanding \emph{why} this asymmetry emerges in learned representations is critical for building systems radiologists can trust.

We adopt the \emph{diagnostic gap} as a task-aware measure of asymmetric prediction stability under perturbation~\cite{icml_short} and use gradient-based latent decomposition of the H-VAE~\cite{sonderby2016ladder,vahdat2020nvae} to explain why this gap widens for fine-grained tasks. 

Our contributions are: (i) a mechanistic explanation of the widened diagnostic gap, showing that weak supervision aligns only a sparse one-dimensional (1D) subspace ($z_1$) while leaving the high-dimensional residual space ($z_{\text{res}}$) structurally vulnerable; (ii) a Hierarchical Variational Autoencoder (H-VAE) with orthogonal latent decomposition that quantitatively separates task-aligned and residual features via analytical gradient-based projection; (iii) a geometric characterization demonstrating that about 95\% of latent capacity is orthogonal to supervision, onsistent with the geometric expectation that alignment of a 1D direction in D-dimensional space scales as $1/\sqrt{D}$~\cite{vershynin2018high}, with empirical scaling $D^{-0.562}$ across $D \in \{32,64,128,256,512\}$ and Stage-2 AUC peaking at $D=128$ (0.574 vs.\ 0.552 at $D=256$); (iv) cross-architecture and cross-modality validation showing a consistent reconstruction stability gap across MIL and MTL on CBIS-DDSM ($\Delta_{\text{diag}} = 1.0\%$--$21.5\%$) and chest X-ray; and (v) statistical validation using $n=1{,}000$ bootstrap confidence intervals, $K=100$ random projection controls, and dimensional sensitivity analysis.

\section{Related Work}

\textbf{Weak Supervision in Medical Imaging.}
MIL and attention-based frameworks~\cite{campanella2019clinical,ilse2018attention} achieve strong classification under limited supervision, but aggregate metrics can obscure task-dependent information loss~\cite{rudin2019stop}. Gradient-weighted Class Activation Mapping (Grad-CAM)~\cite{selvaraju2017gradcam} enables coarse localization from image-level labels and has been applied to weakly supervised lesion detection in mammography~\cite{liu2021weakly}. Prior pipeline-level analyses show that coarse and fine-grained tasks degrade asymmetrically under reconstruction, introducing the diagnostic gap as a task-aware metric~\cite{icml_short}. We provide the first latent-space mechanism explaining this asymmetry.

\textbf{Hierarchical VAEs.}
H-VAEs stack latent levels to model feature hierarchies~\cite{sonderby2016ladder,vahdat2020nvae} and have been applied to medical image synthesis and representation learning~\cite{baur2021autoencoders}. While standard VAEs~\cite{kingma2013auto} and disentanglement variants such as $\beta$-VAE~\cite{higgins2017beta} and FactorVAE~\cite{kim2018factor} promote factor separation, they do not analyze how coarse supervision shapes latent geometry.

\textbf{Latent Space Interpretability and Decomposition.}
Gradient-based methods support saliency and concept attribution~\cite{erhan2009visualizing,simonyan2014deep}, and approaches such as Testing with Concept Activation Vectors, or simply, TCAV map human concepts to linear directions in latent space~\cite{kim2018tcav}. However, these methods are primarily post hoc and do not quantify the proportion of latent capacity governed by supervision. We extend this geometric view by projecting $z$ onto the normalized Stage-1 gradient direction to measure alignment capacity ($\|z_1\|/\|z\|$) and characterizing the orthogonal residual $z_{\text{res}}$ to quantify supervisory sparsity.

\textbf{Simplicity Bias and Shortcut Learning.}
Neural networks exhibit simplicity bias, favoring low-dimensional or shortcut features that minimize loss without ensuring robustness~\cite{geirhos2020shortcut,shah2020pitfalls}. We provide a structural account in generative models: coarse supervision imposes a sparse 1D constraint, leaving fine-grained structure unconstrained in the residual space.

\textbf{Calibration and Reliability.}
Neural networks are often miscalibrated~\cite{guo2017calibration}. Temperature scaling~\cite{guo2017calibration}, Dirichlet calibration~\cite{kull2019dirichlet}, and conformal prediction~\cite{angelopoulos2021gentle} improve reliability. Our focus is representation-level stability; temperature scaling is applied to ensure observed performance gaps are not calibration artifacts.

\section{Method}

We implement an H-VAE to jointly model lesion type~(mass vs.\ calcification) and pathology~(benign vs.\ malignant), analyzing latent feature utilization through gradient-based decomposition.

\subsection{Architecture}

\paragraph{Encoder.}
The encoder $E_\phi$ processes a grayscale ROI 
$x \in \mathbb{R}^{1 \times 256 \times 256}$ 
through an initial convolution followed by four residual downsampling blocks, reducing spatial resolution 
$256 \rightarrow 16$ while expanding channel width ($32 \rightarrow 256$). 
The resulting feature map is flattened and passed through a fully connected layer to produce the parameters 
$\mu, \log\sigma^2 \in \mathbb{R}^{256}$ of the latent distribution. 
Latent samples are drawn via the reparameterization trick~\cite{kingma2013auto}:
\begin{equation}
z = \mu + \sigma \odot \epsilon, \quad \epsilon \sim \mathcal{N}(0, I).
\end{equation}
When decomposition is active, $z$ is reattached to the computation graph to permit gradient-based projection.

\paragraph{Decoder and Classification Heads.}
Three heads branch from the shared latent code 
$z \in \mathbb{R}^{256}$. 
The \textbf{Stage-1 head} is a two-layer fully connected network 
($256 \rightarrow 128 \rightarrow 2$) predicting lesion type (mass vs.\ calcification). 
The \textbf{Stage-2 head} is an identically structured network predicting pathology (benign vs.\ malignant). 
Both heads operate independently on the full $z$. 
The \textbf{reconstruction head} projects $z$ to 
$256 \times 16 \times 16$ via a linear layer, then applies four transposed-convolution blocks to restore resolution to 
$1 \times 256 \times 256$, with a sigmoid output activation.

\paragraph{Training Objective.}
The model is trained by minimising:
\begin{equation}
\mathcal{L} =
\underbrace{\mathcal{L}_{\mathrm{rec}}}_{\text{MSE}}
+ \beta \underbrace{\mathcal{L}_{\mathrm{KL}}}_{\text{KL divergence}}
+ \alpha_1 \underbrace{\mathcal{L}_{c}}_{\text{lesion CE}}
+ \alpha_2 \underbrace{\mathcal{L}_{f}}_{\text{pathology CE}},
\end{equation}
where $\mathcal{L}_{\mathrm{rec}}$ is the mean squared reconstruction error, 
\[
\mathcal{L}_{\mathrm{KL}} = -\frac{1}{2}\sum \left(1 + \log(\sigma^2) - \mu^2 - \sigma^2 \right)
\]
is the KL divergence from $\mathcal{N}(0,I)$, 
and $\mathcal{L}_c$ and $\mathcal{L}_f$ are cross-entropy losses for the Stage-1 and Stage-2 heads respectively.

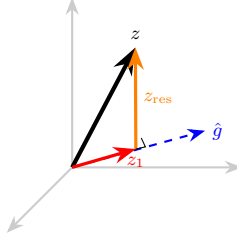
\begin{figure}
  \centering
  \begin{tikzpicture}[>=Stealth, thick, scale=0.75]

    \draw[->, gray!40] (0,0,0) -- (3,0,0);
    \draw[->, gray!40] (0,0,0) -- (0,3,0);
    \draw[->, gray!40] (0,0,0) -- (0,0,3);

    \draw[->, blue, dashed] (0,0,0) -- (2.5,0.8,0.4) 
        node[right, scale=0.8] {$\hat{g}$};

    \draw[->, black, ultra thick] 
        (0,0,0) -- (1.5,2.5,1) 
        node[above, scale=0.8] {$z$};

    \draw[->, red, very thick] 
        (0,0,0) -- (1.2,0.4,0.2) 
        node[below, scale=0.8] {$z_1$};

    \draw[->, orange, very thick] 
        (1.2,0.4,0.2) -- (1.5,2.5,1) 
        node[midway, right, scale=0.8] {$z_{\text{res}}$};

    \draw[black, thin] 
        (1.2,0.4,0.2) 
        -- ++(0.18,0.05,0.03) 
        -- ++(-0.05,0.18,0.06);

  \end{tikzpicture}
  \caption{\small Orthogonal decomposition of $z$ into task-aligned ($z_1$) and residual ($z_{\text{res}}$) components.}
  \label{fig:projection}
\end{figure}

For each sample, we compute the coarse-task gradient 
$g = \frac{\partial h_c(z)_0}{\partial z}$ 
using \texttt{torch.autograd.grad} with \texttt{create\_graph=True}. 
After normalization $\hat{g} = g / (\|g\| + \epsilon)$, $\epsilon=10^{-8}$, 
the latent code is decomposed as:
\begin{equation}
z_1 = \langle z, \hat{g} \rangle \hat{g}, 
\qquad 
z_{\text{res}} = z - z_1.
\end{equation}

By construction, $\langle z_1, z_{\text{res}} \rangle = 0$. 
The component $z_1$ captures variation aligned with the coarse supervisory gradient, 
while $z_{\text{res}}$ contains the orthogonal residual subspace.

This decomposition is analytical and introduces no additional trainable parameters or losses. It enables post hoc analysis of how supervision aligns latent dimensions and how reconstruction affects task-specific stability.

\subsection{Diagnostic Gap Metrics and Evaluation}
\label{sec:gap_metrics}

We adopt the diagnostic gap framework~\cite{icml_short} and apply it within our decomposition to quantify \emph{where} and \emph{why} the gap manifests. Under this framework, a small~$\Delta_{\text{diag}}$ indicates reliable diagnostic preservation under perturbation; a large~$\Delta_{\text{diag}}$ signals task-dependent degradation.

\textbf{Prediction agreement} for task~$\mathcal{T}$ is defined as the fraction of test samples for which the classifier assigns the same predicted class label to the original input~$o$ and the reconstructed input~$r$:
\begin{equation}
\mathrm{Acc}_{\mathcal{T}}^{(o,r)} = \frac{1}{N} \sum_{i=1}^{N} \mathbf{1}\!\left[\hat{y}_{\mathcal{T}}(o_i) = \hat{y}_{\mathcal{T}}(r_i)\right].
\end{equation}
This metric directly measures how much reconstruction alters the classifier's decision, irrespective of ground-truth labels---making it sensitive to representation degradation even when accuracy remains high.

\textbf{Reconstruction stability gap:}
\begin{equation}
\Delta_{\text{diag}} =
\mathrm{Acc}_{\mathcal{T}_c}^{(o,r)} -
\mathrm{Acc}_{\mathcal{T}_f}^{(o,r)},
\label{eq:diagnostic_gap}
\end{equation}
where $\mathcal{T}_c$ is the coarse task (lesion type) and $\mathcal{T}_f$ is the fine-grained task (pathology). Under this framework, $\Delta_{\text{diag}} \in [-1,1]$, since it is the difference between two prediction agreement rates bounded in $[0,1]$. A value near zero indicates comparable stability across tasks (a narrow gap). As $|\Delta_{\text{diag}}|$ increases, the asymmetry widens. A positive gap ($\Delta_{\text{diag}} > 0$) indicates that reconstruction disproportionately degrades the fine-grained task, whereas a negative gap indicates greater degradation of the coarse task. Thus, the magnitude reflects how unequal the degradation is, while the sign indicates which task is more affected.

\textbf{Classification performance gap:} $\Delta_{\text{AUC}} = \mathrm{AUC}_{\mathcal{T}_c} - \mathrm{AUC}_{\mathcal{T}_f}$, capturing inherent discriminative difficulty. Reporting both metrics together distinguishes pre-existing task difficulty ($\Delta_{\text{AUC}}$) from additional degradation introduced by reconstruction ($\Delta_{\text{diag}}$).

\textbf{Classification quality.} Area Under the Receiver Operating Characteristic Curve~(AUC) summarizes discrimination across all thresholds and is the standard metric for imbalanced clinical classification benchmarks~\cite{shen2019deep}. Accuracy is reported as a secondary metric.

\textbf{Calibration.} Expected Calibration Error~(ECE) with temperature scaling~\cite{guo2017calibration} assesses whether confidence scores are reliable, ensuring that observed AUC differences are not artifacts of miscalibration.

\textbf{Supervisory alignment.} The ratio $\|z_1\|/\|z\|$ quantifies what fraction of latent magnitude aligns with the coarse supervisory gradient, enabling geometric comparison to the random-projection baseline.

\section{Experimental Setup}

\subsection{Dataset and Preprocessing}
We evaluate on the CBIS-DDSM dataset~\cite{ddsm} using ROI crops with a patient-wise split: 2,494 training, 536 validation, and 520 test samples (class-balanced via resampling for mass/calcification and benign/malignant). Images were resized to $256{\times}256$ and intensity-normalized to $[0,1]$.

\subsection{Training Protocol}

Models were trained end-to-end with the Adam optimizer (learning rate $10^{-4}$, weight decay $10^{-5}$), batch size 32, and a maximum of 30 epochs. We report results for $\beta=0.05$ ($\alpha_1 = \alpha_2 = 1.0$). All weights were initialized with Xavier normal initialization~\cite{glorot2010understanding}; gradient clipping (max norm~1.0) and early stopping (patience~5, monitoring validation reconstruction loss) were applied.
We report main results for $D=256$ to maintain parity with standard generative baselines, though dimensional ablations across $D \in \{32, 64, 128, 256, 512\}$ are shown in Table \ref{tab:dim_main}.

\section{Results}

\subsection{Classification and Diagnostic Gaps}

The right side of Table~\ref{tab:dim_main} summarizes classification performance on CBIS-DDSM. 
Stage-1 substantially outperforms Stage-2, yielding a large classification gap 
($\Delta_{\text{AUC}} = 0.314$, $p < 0.001$). Temperature scaling reduces ECE 
for both stages, confirming that this asymmetry is not attributable to miscalibration.

Reconstruction further amplifies this disparity. Stage-1 prediction agreement 
exceeds Stage-2 agreement, producing a significant reconstruction stability gap 
($\Delta_{\text{diag}} = 0.050$, 95\% CI: 0.012--0.083; $p = 0.005$). 
Thus, the fine-grained task is both intrinsically harder and more sensitive 
to reconstruction perturbation.

\begin{table}[t]
\centering
\caption{Left: Dimensional sensitivity analysis across $D \in \{32,\ldots,512\}$; 
Right: Classification and reconstruction stability on CBIS-DDSM (test set); 
bootstrap 95\% CIs ($n=1{,}000$), ECE after temperature scaling.}
\label{tab:dim_main}

\begin{minipage}[t]{0.54\textwidth}
\centering
\begin{tabular}{lcccc}
\toprule
$D$ & S1 AUC & S2 AUC & $\Delta_{\text{diag}}$ & $\|z_1\|/\|z\|$ \\
\midrule
32  & 0.799 & 0.562 & 0.038 & 0.165 \\
64  & 0.848 & 0.498 & 0.077 & 0.080 \\
128 & 0.873 & 0.574 & 0.044 & 0.087 \\
256 & 0.866 & 0.552 & 0.050 & 0.039 \\
512 & 0.868 & 0.563 & 0.052 & 0.034 \\
\bottomrule
\end{tabular}
\end{minipage}
\hfill
\begin{minipage}[t]{0.42\textwidth}
\centering
\begin{threeparttable}
\begin{tabular}{lcccc}
\toprule
Stage & AUC & 95\% CI & ECE & Agree. \\
\midrule
Stage-1 & 0.866 & (0.835, 0.895) & 0.129 & 92.69\% \\
Stage-2 & 0.552 & (0.500, 0.602) & 0.305 & 87.69\% \\
\bottomrule
\end{tabular}
\begin{tablenotes}
\small
\item \textit{Note:} As $D$ increases, the sparsity of the supervisory signal 
($\|z_1\|/\|z\|$) becomes more pronounced, corresponding to an increase in the 
diagnostic gap from 0.038 to 0.052.
\end{tablenotes}
\end{threeparttable}
\end{minipage}

\end{table}

\subsection{Latent Geometry and Ablation}

Only $\approx$4.4\% of latent magnitude aligns with the Stage-1 gradient 
(95\% CI: 4.1--4.7\%), with $|\cos(z_1, z_{\text{res}})| < 10^{-4}$ confirming 
orthogonal separation. This is consistent with the geometric expectation 
$1/\sqrt{256} = 6.25\%$ for a random 1D vector~\cite{vershynin2018high} ($p=0.660$), 
indicating that sparsity is an intrinsic property of high-dimensional space, not 
a training failure. Yet the gradient direction retains functional signal: Stage-1 
AUC from $z_1$ (0.611) exceeds the random projection mean ($0.525 \pm 0.089$), 
suggesting it captures task-relevant structure ($p=0.160$; see Table~\ref{tab:dim_main}).

Latent ablation confirms that both tasks depend primarily on $z_{\text{res}}$: 
removing it degrades both stages to near-random performance, while removing $z_1$ 
preserves near-full capability. The 1D gradient direction lacks the capacity for 
full discrimination, structurally forcing both tasks into the residual subspace, 
where fine-grained pathology is disproportionately affected due to its greater difficulty.

\begin{table}[t]
\centering
\caption{Left: Classification AUC under latent ablation. Right: Cross-modality validation.}
\label{tab:ablation_generalization}
\begin{minipage}[t]{0.44\textwidth}
\centering
\begin{tabular}{lcc}
\toprule
Latent Input & S1 AUC & S2 AUC \\
\midrule
Full $z$         & 0.866 & 0.552 \\
$z_1$ only       & 0.611 & 0.523 \\
$z_{\text{res}}$ only & 0.801 & 0.538 \\
\bottomrule
\end{tabular}
\end{minipage}
\hfill
\begin{minipage}[t]{0.52\textwidth}
\centering
\begin{tabular}{lccc}
\toprule
Dataset & S1 AUC & S2 AUC & $\Delta_{\text{diag}}$ \\
\midrule
CBIS-DDSM  & 0.866 & 0.552 & 0.050 \\
Chest X-Ray & 0.986 & 0.820 & 0.056 \\
\bottomrule
\end{tabular}
\end{minipage}
\end{table}
\subsection{Generalization and Method Comparison}

The diagnostic gap persists across modalities. To validate cross-domain generalization, we evaluated the H-VAE framework on a pediatric Chest X-ray dataset~\cite{chestkaggle}, defining the coarse task as pneumonia detection (normal vs.\ pneumonia) and the fine-grained task as etiology classification (bacterial vs.\ viral). Stage-1 AUC demonstrates near-perfect classification performance (0.986) while Stage-2 AUC drops to 0.820, yielding a classification gap of 0.166. Reconstruction stability shows a consistent asymmetry ($\Delta_{\text{diag}} = 0.056$), confirming that the diagnostic gap generalizes beyond mammography to other hierarchical medical imaging tasks.

Across weakly supervised baselines on CBIS-DDSM, perturbation structure strongly influences gap magnitude. Input-space perturbations like instance dropout in MIL produce a narrow gap (0.010), while latent-space perturbations expose structural vulnerabilities: reconstruction in H-VAE yields a moderate gap (0.050), and isotropic Gaussian noise in MTL produces a substantially wider gap (0.215) by uniformly disrupting the unprotected orthogonal dimensions.

\section{Discussion}
\label{sec:discussion}

Gradient-based decomposition confirms that coarse supervision is confined to a sparse 1D latent direction ($z_1$, $\approx$4.4\% of magnitude), yet ablation shows this dimension alone cannot discriminate either task. Both coarse and fine-grained features are therefore forced into $z_{\text{res}}$, where fine-grained pathology bears a greater representational burden, providing a natural explanation for the \emph{diagnostic gap}.

\textbf{Perturbation dependence.} The magnitude of $\Delta_{\text{diag}}$ varies with perturbation type. Localized perturbations (e.g., instance dropout) yield narrow gaps, while uniform perturbations (e.g., Gaussian noise) widen the gap by disrupting orthogonal latent dimensions. The positive diagnostic gap on chest X-ray demonstrates that this asymmetric degradation persists even in a near-ceiling coarse-task regime, confirming that perturbation structure and task hierarchies jointly determine gap magnitude.

\textbf{Calibration.} While post-hoc calibration methods improve predictive confidence, our analysis targets representation-level stability rather than uncertainty estimation.

\textbf{Architectural implications.} Compared to a standard VAE, our H-VAE approach highlights where architectural vulnerabilities lie. Explicit geometric regularization that expands the task-aligned subspace may be necessary to improve fine-grained robustness. Empirically, the $D=128$ latent dimensionality provides the best trade-off between stability and compactness (see Table~\ref{tab:dim_main}).

\section{Conclusion}

We presented an H-VAE with gradient-based latent decomposition that provides a mechanistic explanation for the widening diagnostic gap under weak supervision. We show that coarse supervision inherently targets only a geometrically sparse 1D latent subspace, structurally forcing both coarse and fine-grained pathology features to depend heavily on the high-dimensional orthogonal residual. This geometric reality explains why fine-grained stability is disproportionately exposed to perturbation, and why the \emph{gap} exists.

Future work will extend this framework to multi-label and hierarchical supervision, explore transformer-based encoders where gradient directions may be more distributed, and use the geometric characterization to design supervision strategies that deliberately expand the constrained subspace to improve fine-grained clinical reliability.

\section*{Impact Statement}
This work improves the transparency of weakly supervised clinical screening. By providing a structured account of feature degradation, our framework identifies when models rely on shortcuts rather than robust pathological cues. This supports safer deployment of diagnostic pipelines by offering a rigorous standard beyond aggregate AUC. While reducing annotation burdens, we emphasize that these systems should be used as supportive tools alongside radiologist oversight.

\bibliographystyle{splncs04}
\bibliography{mybibliography}

\end{document}